# AI Generations: From AI 1.0 to AI 4.0

Jiahao Wu[1], Hengxu You[2], Jing Du, Ph.D.[3*]

[1]Ph.D. Student, Informatics, Cobots and Intelligent Construction (ICIC) Lab, Engineering School of Sustainable Infrastructure & Environment, University of Florida, Gainesville, FL 32611; e-mail: jiahaowu@ufl.edu

[2]Ph.D. Candidate, Informatics, Cobots and Intelligent Construction (ICIC) Lab, Engineering School of Sustainable Infrastructure & Environment, University of Florida, Gainesville, FL 32611; e-mail: you.h@ufl.edu

[3]Professor, Informatics, Cobots, and Intelligent Construction (ICIC) Lab, Engineering School of Sustainable Infrastructure & Environment, Department of Mechanical & Aerospace Engineering, Department of Industrial & System Engineering, University of Florida, Gainesville, FL 32611 (corresponding author); Email: eric.du@essie.ufl.edu

**ABSTRACT**

This paper proposes that Artificial Intelligence (AI) progresses through several overlapping generations: AI 1.0 (Information AI), AI 2.0 (Agentic AI), AI 3.0 (Physical AI), and now a speculative AI 4.0 (Conscious AI). Each of these AI generations is driven by shifting priorities among algorithms, computing power, and data. AI 1.0 ushered in breakthroughs in pattern recognition and information processing, fueling advances in computer vision, natural language processing, and recommendation systems. AI 2.0 built on these foundations through real-time decision-making in digital environments, leveraging reinforcement learning and adaptive planning for agentic AI applications. AI 3.0 extended intelligence into physical contexts, integrating robotics, autonomous vehicles, and sensor-fused control systems to act in uncertain real-world settings. Building on these developments, AI 4.0 puts forward the bold vision of self-directed AI capable of setting its own goals, orchestrating complex training regimens, and possibly exhibiting elements of machine consciousness. This paper traces the historical foundations of AI across roughly seventy years, mapping how changes in technological bottlenecks from algorithmic innovation to high-performance computing to specialized data, have spurred each generational leap. It further highlights the ongoing synergies among AI 1.0, 2.0, 3.0, and 4.0, and explores the profound ethical, regulatory, and philosophical challenges that arise when artificial systems approach (or aspire to) human-like autonomy. Ultimately, understanding these evolutions and their interdependencies is pivotal for guiding future research, crafting responsible governance, and ensuring that AI's transformative potential benefits society as a whole.

## I. INTRODUCTION

Artificial Intelligence (AI) has experienced a transformative evolution over the last seventy years, evolving from its nascent stage of theoretical formulations to its current status as a cornerstone of technological advancement [1]. Initially, the field was dominated by intellectual explorations into symbolic reasoning, knowledge representation, and the rudimentary principles of machine learning [2]. These early stages were marked by a focus on conceptual breakthroughs, laying the groundwork for what AI could potentially achieve. As computational capabilities expanded and data sources proliferated, AI transitioned from theoretical models to practical applications capable of learning from patterns and making precise predictions [3]. The last two decades, however, have witnessed an unprecedented acceleration in AI development, propelling the field into realms that surpass even the most optimistic projections of its early pioneers.

Despite remarkable successes in areas like natural language processing, computer vision, and large-scale data analytics, AI continues to face challenges in interacting seamlessly with complex, dynamic real-world environments. This ongoing struggle signals an emerging phase in AI's evolution, marking a shift from systems that primarily process and predict information to ones that can plan, decide, and act, ushering in new generations of AI: *Information AI (AI 1.0)*, *Agentic AI (AI 2.0)*, *Physical AI (AI 3.0)* and *Conscious AI (AI 4.0)*. This classification not only clarifies the conceptual transitions within the field but also helps delineate the evolution of AI capabilities from data extraction to making autonomous decisions in digital realms, and now to engaging directly with the physical world.

Understanding these transitions is essential, not just from a technological standpoint but also for grasping the societal and economic implications of AI. Each phase of AI has been shaped by distinct technological drivers and bottlenecks: the early period was limited by the lack of advanced algorithms and computational frameworks [4]; the advent of powerful GPUs around 2012 significantly shifted the landscape, enabling more complex neural architectures [5]; and today, the challenge has moved towards harnessing domain-specific, high-quality data to feed into these sophisticated systems [6]. Recognizing these shifts is crucial for stakeholders, including policymakers, researchers, and industry leaders, who must navigate the ethical, regulatory, and technical complexities introduced by advanced AI systems.

The objective of this review is to provide a comprehensive retrospective on the milestones that have defined AI's progress. By tracing the lineage of algorithmic innovations, increases in computing power, and enhancements in data utilization, we aim to illuminate the significant moments that have shaped AI from its inception to its current state. This exploration is structured around the AI 1.0 to AI 4.0 framework, illustrating how each generation's defining features and limitations correspond to broader historical phases from approximately 1950 to the present. In doing so, we will also contemplate the future trajectory of AI, considering the potential technical challenges, societal impacts, and strategic directions that could define the next phases of AI research and application.

This article is structured to first revisit the historical foundations of AI, emphasizing the shifts in primary drivers from algorithms to computing power to data. We then delve into the specific characteristics, achievements, and limitations of AI 1.0, AI 2.0, AI 3.0, and AI 4.0. Following this, we explore the convergence and future outlook of AI, highlighting the synergies among the four generations and outlining the grand challenges that lie ahead. Finally, we conclude with a synthesis of key insights and propose future directions for sustained progress in the field, aiming to both inform and inspire continued innovation and thoughtful integration of AI into our daily lives and societal structures.

## II. HISTORICAL FOUNDATIONS OF AI

*2.1 Phase 1 (1950s-2010s): Age of Algorithmic Innovations*
Since the 1950s, AI has advanced through a dynamic interplay among three core ingredients: *algorithms*, *computing power*, and *data* [7]. Although these three factors have always shaped the field, they have not always contributed equally at every stage. In the early decades, the limiting factor was innovation in algorithms. From mid-century debates about the feasibility of machine intelligence to the emergence of expert systems and neural networks, it was clear that conceptual breakthroughs would determine AI's boundaries [8]. Meanwhile, although data and computing power were important, they played more supportive roles. Gradually, as new hardware architectures appeared and as large-scale datasets became more accessible, the focus shifted toward harnessing immense computational capability and vast amounts of information.

From the outset, researchers were enthralled by the question of whether machines could truly think. Alan Turing's pioneering paper [9] set the stage, posing the famous "imitation game" as a litmus test for intelligence. In 1956, the Dartmouth Conference [10] formally introduced the term "Artificial Intelligence" and laid out the bold proposition that the essence of human intelligence could be precisely described and replicated in machines. Early systems, such as the *Logic Theorist* and the *General Problem Solver* [2,11] underscored that symbolic reasoning could be computationally realized. These proof-of-concept attempts highlighted the central premise of that era: if we could devise the right algorithms, computers might reason and solve problems with near-human efficacy.

By the 1960s and 1970s, a strong emphasis on *symbolic AI* took hold. Influential works by John McCarthy [12] introduced LISP as a language suited to symbolic processing, while Minsky and Papert's [13] critical analysis of single-layer perceptions contributed to a pause in neural network research, pushing many researchers toward knowledge-based or "expert" systems. Milestones like the DENDRAL project [14] and MYCIN [15] showcased how carefully curated rule sets could guide problem-solving in specialized domains. These systems illustrated the power of algorithmic design in areas such as medical diagnosis or chemical analysis, even when real-world data were scarce and computational resources limited.

Neural networks rebounded in the 1980s with work on Hopfield networks [16] (**Fig.1**) and, crucially, the rediscovery of backpropagation [17]. This gave researchers fresh insight into how machines might learn patterns from data. Though the potential of these connectionist approaches was clear, they often stalled because large datasets were not widely available and specialized hardware did not yet exist. Even so, foundational contributions like LeCun, et al. [18] application of *convolutional neural networks* to handwritten digit recognition laid the groundwork for what would become modern deep learning.

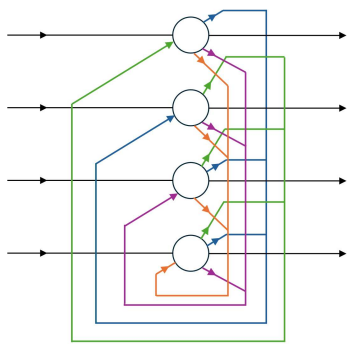

**Fig.1** The Hopfield networks [16] introduced content-addressable memory in neural networks, marking a major milestone in connectionism in AI.

By the 1990s, certain algorithmic achievements hinted at deeper architectures capable of tackling increasingly complex tasks. The proposal of Long Short-Term Memory (LSTM) networks effectively addressed the vanishing gradient problem, opening possibilities for modeling sequential data more accurately [19]. However, the real transformative moment emerged around 2012, when Krizhevsky, Sutskever, and Hinton demonstrated that ImageNet-scale datasets and high-performance GPUs could dramatically improve a deep neural network's ability to classify images, i.e., the *AlexNet* [20] (**Fig.2**). Although this watershed event is often viewed as the dawn of the "deep learning era," it could not have happened without the algorithmic groundwork laid over the preceding decades.

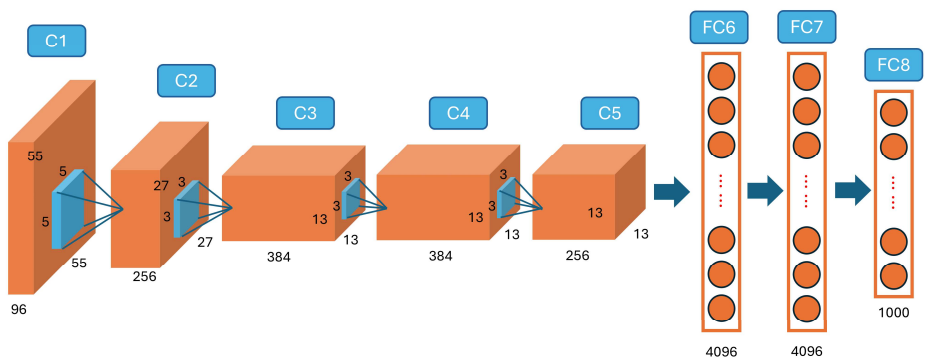

**Fig. 2** AlexNet [20] marks the beginning of large-scale, GPU-accelerated convolutional neural networks for high-performance image classification

*2.2 Phase 2 (2010s- Present): The Compute Revolution and Deep Learning Renaissance*
A dramatic shift in AI research took hold around 2012, when mounting computational capacity began to eclipse algorithmic novelty as the principal engine of progress. While the core concepts underlying neural networks had been present since at least the 1980s, it was the widespread adoption of General-Purpose Graphics Processing Units (GPUs) that ignited what is often termed the "*deep learning renaissance*" (**Fig.3**). When Krizhevsky, et al. [20] leveraged GPUs to train a large convolutional neural network for the ImageNet competition, they decisively demonstrated how parallelized computing could unearth performance gains previously unachievable with single-threaded Central Processing Units (CPUs). This turning point catalyzed a wave of research across machine vision, speech recognition, and natural language processing, with groups at Google, Microsoft, Baidu, and many academic institutions all racing to scale up network architectures [21-23]. The essence of this period lay in the conviction that "bigger is better", whether in terms of model parameters, dataset size, or sheer computational resources. Consequently, much of the state-of-the-art progress hinged on harnessing specialized hardware: first GPUs, then tensor processing units (TPUs) and other custom accelerators, to churn through ever-growing datasets in shorter training cycles.

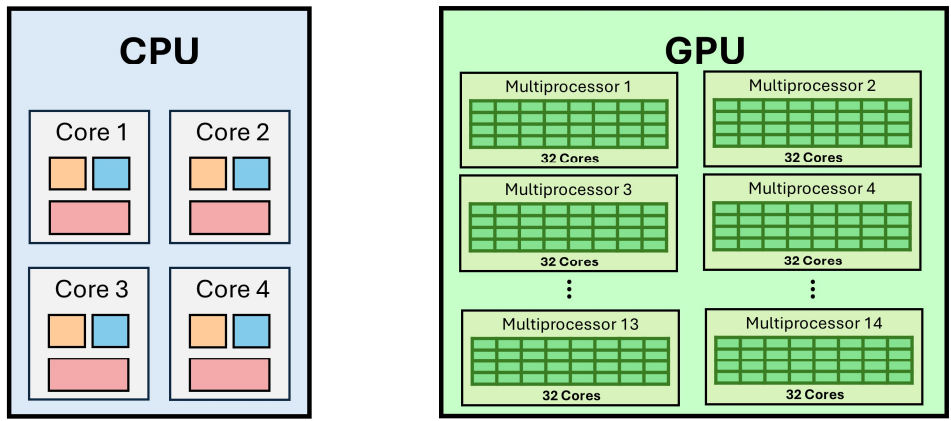

**Fig.3** The CUDA architecture pioneered general-purpose GPU computing, revolutionizing parallel processing and accelerating AI breakthroughs.

By the mid-2010s, the explosive rise of deep reinforcement learning [24] and breakthroughs in game-playing AI, such as AlphaGo [25], underscored that not only could AI models learn representations from massive data, but they could also discover winning strategies through large-scale simulations. Nevertheless, the predominant realm for these systems remained resolutely digital. Whether classifying images, translating text [26,27], or playing complex board and video games, AI was still operating in an essentially informational context. Although data availability was critical and algorithms like convolutional and recurrent neural networks continued to improve, sheer computational power was often the deciding factor in achieving superior performance. Researchers observed emergent patterns in scaling laws[28], revealing that larger models trained on larger datasets could unlock qualitatively new capabilities. Systems like GPT-2 [29] and GPT-3 [30] illustrated this phenomenon vividly by demonstrating a striking ability to generate human-like text once parameter counts and training data reached certain thresholds. For all their sophistication, these models continued to reside in the digital world, making them refined and powerful versions focused on big data analytics and pattern recognition at an unprecedented scale. Even so, the end of this phase began to hint at a transition toward greater autonomy and decision-making in digital contexts, an emerging hallmark of agentic AI. While many systems are still centered on classification or prediction, the rise of advanced reinforcement learning agents able to adapt strategies within software ecosystems foreshadowed a new kind of agency. By approximately 2024, the scholarly and commercial drive to develop goal-directed virtual assistants, automated resource allocation tools, and multi-agent simulations suggested that the chief challenge was no longer purely to label data accurately, but to act in digital environments in ways that transcended traditional supervised learning [31]. This growing desire for agentic AI remained tied to abundant computing power, yet it began to reveal new dependencies on specialized data streams and real-time feedback loops [32]. It set the stage for the next generation of AI, in which computational needs would remain vital, but data and context-specific knowledge would become even more pivotal in enabling truly autonomous, adaptive systems.

*2.3 Phase 3 (2024 – Foreseeable future): Data-Centric Paradigms*
In the wake of a period defined by dramatic increases in computational horsepower, the focal point of AI advancement has shifted once again. Where Phase 2 thrived on scaling neural networks through unprecedented parallel processing, Phase 3 acknowledges that data, especially specialized, high-quality data, is frequently the greatest obstacle. Researchers have discovered that ever-larger models alone do not guarantee success if they lack context-rich training sets. Consequently, a surge in large-scale, domain-specific data-collection efforts has emerged, reshaping the field's priorities. Projects that aggregate specialized medical data for diagnostic systems [33], simulate high-fidelity environments for robotics and autonomous vehicles [34,35], or compile deep reinforcement learning benchmarks with realistic constraints [36,37] attest to the idea that harnessing robust datasets can be as determinative as algorithmic ingenuity or raw computational power.

Despite the continued importance of parallel computing and innovative architectures, many cutting-edge successes now hinge on data strategy. Researchers have championed "data-centric AI" [38], arguing that refining training sets, removing biases, filling in coverage gaps, or generating synthetic data to handle edge cases, often yields more improvement than adding layers to a neural network. This philosophy is closely related to the rise of foundation models [39], which are vast neural architectures that can be adapted to myriad tasks, but require massive, carefully curated corpora to realize their full potential. As data becomes the true bottleneck, teams must grapple with the logistical and ethical challenges of collecting, storing, and labeling it, as well as with privacy, consent, and representation issues.

Within this phase, AI's transition from informational analysis to agentic decision-making becomes increasingly tangible. Reinforcement learning agents not only plan and learn in complex digital worlds but also begin to bridge into real-world applications, where they must reason about noisy sensors, hardware uncertainties, and human collaboration. Physical AI, exemplified by advanced robotics, autonomous drones, and integrated cyber-physical systems, moves beyond the boundaries of simulated or purely informational spaces. Progress in robotic grasping and manipulation [35,40], self-driving vehicles [41], and robotic surgery [42] signals how these systems can robustly interact with the environment, handle dynamic conditions, and learn from continuous feedback. Thus, the hallmark of this new phase is the recognition that data unlocks the fuller potential of agentic AI in digital ecosystems, as well as physically embodied intelligence in the real world [43].

### III. AI GENERATIONS

The historical review of AI underscores a pivotal generational shift and evolution in AI paradigms, calling for a novel framework for understanding and classifying AI. In this context, we avoid the traditional technical definitions that categorize AI strictly by their operational or algorithmic characteristics. Instead, our analysis seeks to understand AI through its intrinsic qualities: *What are they? What are they designed to achieve?* And *what are their consistent behavioral patterns?* Accordingly, we propose a taxonomy that identifies four distinct generations of AI: AI 1.0, characterized as Information AI, which focuses on data processing and knowledge management; AI 2.0, or Agentic AI, which encompasses systems capable of autonomous decision-making; AI 3.0, known as Physical AI, which integrates AI into physical tasks through robotics; and the speculative AI 4.0, termed Conscious AI, which posits the potential emergence of self-aware AI systems. This classification aims to provide a more detailed perspective reflecting AI technologies' complex evolution. **Fig.4** illustrates the generational evolution of artificial intelligence (AI) from AI 1.0 (Information AI) to AI 4.0 (Conscious AI).

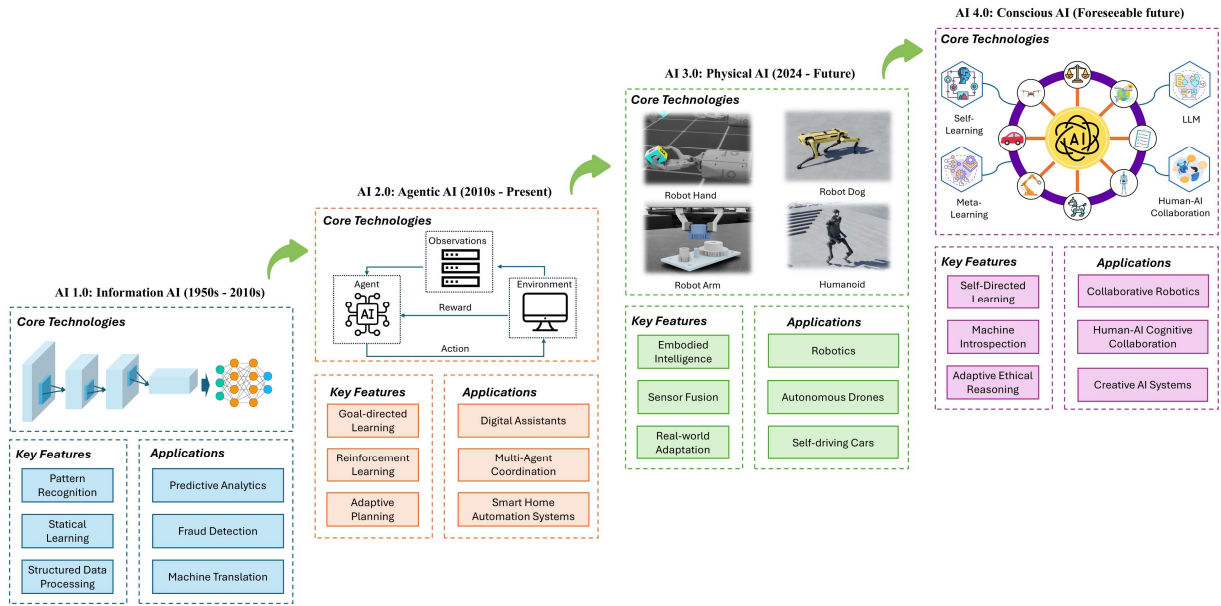

**Fig. 4** The Evolution of AI Generations from AI 1.0 to AI 4.0

*3.1 AI 1.0: Information AI*

The concept of AI 1.0 captures a stage in which computational systems excel at classifying and interpreting information but remain confined to analyses of static data, rather than engaging in active decision-making or real-world manipulation. Fundamentally, AI 1.0 focuses on pattern recognition and information processing, techniques that have powered breakthroughs in computer vision, natural language processing (NLP), and recommendation systems. Although these achievements might seem commonplace now, they represent the fruits of decades of research driven by both mathematical innovation and the increasing availability of digital data.

Many of the core ideas underpinning AI 1.0 trace back to early neural network research and statistical machine learning. From Rosenblatt's perceptron in the late 1950s to the backpropagation algorithms popularized by Rumelhart, Hinton, and Williams [17], these developments laid the groundwork for data-driven learning by demonstrating that machines could uncover patterns within examples rather than relying solely on hand-coded rules. Classic approaches to supervised learning, such as Support Vector Machines (SVMs) formalized by Cortes and Vapnik [44], later proved to be formidable contenders in tasks ranging from handwriting recognition to text classification. Progress in computational hardware, along with the accumulation of sizeable labeled datasets, eventually made it feasible to train deeper and more complex neural networks, culminating in milestone successes in computer vision. A watershed moment came when Krizhevsky et al. [20]'s AlexNet leveraged parallelized GPU training to conquer the ImageNet challenge, revealing how convolutional architectures could outperform all prior methods by learning increasingly abstract features from raw image pixels.

In natural language processing, the influence of AI 1.0 can be seen in early sequence models and statistical language modeling. Although these systems often relied on simpler Markov or n-gram assumptions, they set the stage for more advanced architectures by highlighting the necessity of abundant text corpora. Meanwhile, recommendation engines, such as those popularized by the Netflix Prize [45], underscored how analyzing large-scale user interactions could drive consumer engagement on streaming and e-commerce platforms. Today, many companies still rely on these core AI 1.0 technologies, sometimes enhanced with shallow neural architectures, to filter spam, rank search results, recommend products, or detect fraudulent transactions. Indeed, for structured or semi-structured data, these pattern-recognition approaches remain both cost-effective and highly accurate.

Despite their deep societal impact, AI 1.0 systems generally lack autonomy or contextual awareness associated with subsequent generations of AI. They excel at predicting outcomes when provided with substantial training data, but they require a relatively stable environment and benefit most from human supervision in data curation and decision-making. Performance often degrades if the input distribution shifts significantly, a vulnerability illustrated when face recognition models falter on underrepresented groups or when language models encounter domain-specific jargon. While the considerable success of AI 1.0 is undeniable - transforming industries from finance to healthcare through improved analytics and diagnostics - its limitations lie in its reactive nature. Pattern recognition alone offers no guarantee of proactive decision-making, real-time adaptation, or safe deployment in dynamic settings. These constraints, while hardly trivial, became the springboard for further developments in AI 2.0 and 3.0, in which systems aim to learn, plan, and even act within uncertain digital or physical worlds.

*3.2 AI 2.0: Agentic AI*
A defining characteristic of AI 2.0 is the emergence of systems capable of autonomous decision-making within digital contexts. Rather than merely classifying static data, these agents adapt their behavior to achieve goals, often in complex or continuously evolving environments. Reinforcement learning (RL) has played a pivotal role in this shift, enabling machines to learn strategies by interacting with simulated or real-world settings and receiving feedback in the form of rewards or penalties. Pioneering work on deep RL [24] and subsequent achievements such as AlphaGo [25] underscored how sufficiently powerful algorithms and ample computing resources could surpass human performance in tasks that demand long-term planning and strategic adaptation. A common thread among these systems is the concept of goal-directed planning: software agents allocate resources, schedule tasks, or coordinate with other agents, leveraging sophisticated RL or hybrid RL-language model algorithms [30] that integrates contextual understanding (**Fig.5**).

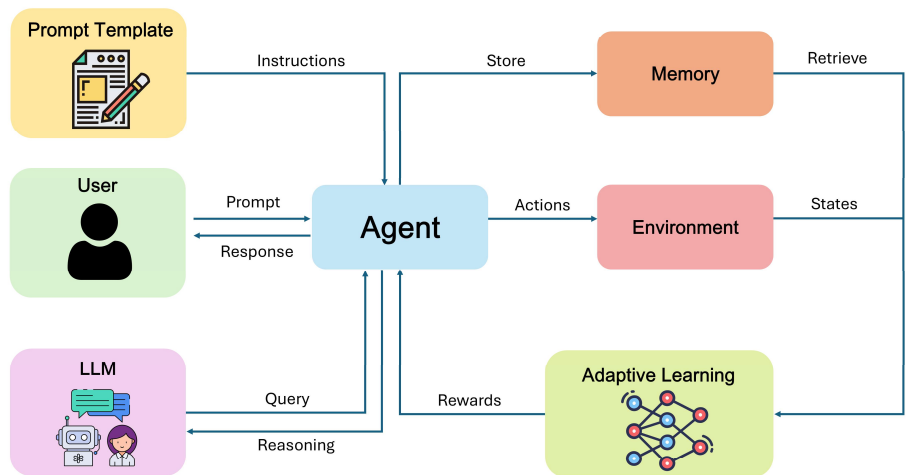

**Fig.5** Agentic AI uses adaptive policies, enabling autonomous action and continuous self-improvement.

Although the conceptual leap from AI 1.0's pattern recognition to AI 2.0's agentic behavior might appear seamless, it demands a unique confluence of technical elements. Computing power is crucial because agentic systems frequently require real-time inference and the ability to run complex simulations, whether they involve a marketplace, a multiplayer environment, or the robust scheduling of cloud resources. The pursuit of these computationally intensive tasks has spurred the development of GPU clusters, tensor processing units (TPUs), and other specialized accelerators designed for iterative training and low-latency decision-making.

Alongside raw computing, data now shifts toward contextual, time-varying inputs. Instead of static image sets, these systems often ingest streams of logs, market quotes, event triggers, or user interactions.

Training an agent to trade stocks automatically or to operate a recommendation engine in real-time requires ongoing ingestion of behavioral data and a capacity to adapt as market conditions or user preferences evolve. In parallel, algorithms for planning and multi-agent coordination continue to mature. RL frameworks have grown more refined, incorporating hierarchical strategies [46], policy optimization methods [47], and combinations with large language models to generate more adaptive and context-aware decisions.

Practical applications of AI 2.0 already abound, even if many are not labeled "reinforcement learning" by name. Automated trading systems in finance exemplify how agents make high-frequency decisions under uncertainty, guided by streaming data feeds. Recommendation systems, evolving from static collaborative filtering, increasingly incorporate feedback loops to adapt suggestions in real time, improving user engagement across e-commerce and media platforms. Digital assistants and software schedulers, while not yet ubiquitously agentic, offer glimpses of a future where AI handles tasks like resource allocation, task delegation, and multi-agent coordination within corporate or consumer software ecosystems. Projects showcasing multi-user environment simulations, such as AI-driven group scheduling bots, complex traffic simulations, or large-scale online game AI [48], further illustrate how these agentic systems anticipate and respond to dynamic conditions.

Viewed from a societal vantage, AI 2.0 promises efficiency gains in many sectors, ranging from manufacturing pipelines that automatically schedule production runs to logistics networks that allocate trucks or drones in real time. Nonetheless, expanded autonomy introduces ethical and policy dilemmas. When decisions are made algorithmically, issues of bias, privacy, and accountability become magnified. Consider an agentic recommendation engine that adapts its suggestions to maximize user "clicks" or "watch time": if left unchecked, such optimization can exacerbate echo chambers or inadvertently spread disinformation. Similarly, automated trading agents may destabilize financial markets if they act on unforeseen correlations or maladaptive reward incentives. The challenge, therefore, lies in ensuring that the computational, data-centric, and algorithmic foundations of AI 2.0 are harnessed responsibly. In the push toward future AI systems, balancing autonomy with transparency and fairness will be as crucial to societal acceptance as any technical advancement.

*3.3 AI 3.0: Physical AI*

Where AI 1.0 has excelled in analyzing data and AI 2.0 in making decisions within digital realms, AI 3.0 takes intelligence off the screen and into the physical world. At its core, this phase is defined by embodied systems that perceive, plan, and act in real time under conditions of uncertainty and complexity. Fields like robotics, autonomous vehicles**,** drones, industrial automation, **and** surgical robotics have become the living laboratories of AI 3.0, integrating machine learning with mechanical and electronic control systems. The unifying characteristic is that these intelligent agents no longer remain passive observers or purely virtual actors; instead, they directly sense their environment through arrays of sensors and respond through actuators that exert forces, move limbs, or navigate terrains.

A central challenge in bringing physical AI to life lies in data acquisition. Unlike digital contexts where data can be abundant and neatly labeled, physical systems demand high-fidelity sensor data that accurately represents an environment's complexity, from variable lighting conditions to changing weather patterns. This need for domain-specific, robust data complicates design and training. A robot operating on a factory floor requires carefully calibrated cameras, LiDAR, or haptic sensors, while an autonomous drone might rely on GPS, inertial measurement units, and computer vision to navigate. Each sensor stream demands real-time processing and reliable fusion techniques to provide a coherent view of the world. Consequently, computing power in AI 3.0 shifts towards distributed and edge computing architectures. Systems must often process sensor inputs on-board to make split-second decisions, i.e., an imperative that underscores the importance of energy-efficient hardware, specialized accelerators, and potentially 5G or 6G networks that reduce communication latency when data must be shared with cloud resources.

On the algorithmic front, physical AI blends advanced machine learning with control theory and systems engineering. RL has demonstrated promise in tasks like robotic grasping and manipulation [35,40], but real-world settings introduce complexities such as partial observability, unpredictable disturbances, and the need for robust or safe RL strategies [49]. Sophisticated sensor fusion methods [50] are essential for integrating heterogeneous sensor inputs, while advanced control techniques [51,52] ensure that autonomous vehicles and robots can move fluidly and interact safely with humans. Designing systems that gracefully handle failures or anomalies, such as a malfunctioning sensor or unforeseen obstacles, further emphasizes the importance of redundancy and resilience in both hardware and software.

The real-world impact of AI 3.0 is already evident across multiple domains. In manufacturing, co-robots work collaboratively on assembly lines, lifting heavy parts or performing precision tasks, drastically reducing workplace injuries and boosting productivity. In healthcare, semi-autonomous surgical systems [42] enable finer control in minimally invasive procedures, while eldercare robots assist with daily activities in retirement communities. Construction and logistics industries are also adopting autonomous machinery and robotic fleets to optimize workflows and reduce labor costs. These trends benefit from an increasing intersection with the Internet of Things (IoT) and next-generation connectivity (5G/6G), forging cyber-physical systems in which objects, sensors, and AI agents coordinate to improve efficiency and safety.

However, the leap from digital to physical deployment exposes AI to a new realm of uncertainties. Environmental extremes, unstructured terrain, or the unpredictability of human interactions pose significant risks. Even small design oversights can have dire consequences when a physically embodied system malfunctions, such as a self-driving car encountering sudden obstacles [41] or a warehouse robot navigating crowded aisles. Safety, reliability, and regulatory compliance thus loom as major challenges, prompting debates over liability if accidents occur. Setting standards for autonomous driving (NHTSA guidelines, ISO 26262 for functional safety in road vehicles) or robot operation in human-centric environments becomes paramount to public acceptance. The question of ethical deployment extends further still: as drones or industrial robots proliferate, policymakers, manufacturers, and citizens must grapple with the implications for labor markets, data privacy, and environmental impact.

### *3.4 AI 4.0: Conscious AI*

The notion of AI 4.0 envisions systems that go beyond the ability to interpret information (AI 1.0), act in digital contexts (AI 2.0), or react to the physical world (AI 3.0). Instead, these hypothetical agents would set their own goals, comprehend environments (whether digital, physical, or hybrid), and train and orchestrate themselves (including selecting and combining multiple models) without human intervention. Proponents of this idea contend that once AI systems acquire sufficient complexity and sophistication, they may exhibit forms of machine consciousness comparable to human subjective experience or self-awareness [53]. Although this is a bold and highly controversial claim, it underscores a growing conversation about the final frontiers of intelligence and autonomy.

A key challenge in discussing conscious AI arises from the fact that no universally accepted definition or theory of consciousness exists, even among neuroscientists, cognitive scientists, and philosophers of mind. Some theorists ground consciousness in information integration and complexity, as in Tononi's Integrated Information Theory [54,55], while others emphasize global workspace architectures [56,57]. Philosophers like David Chalmers [58] frame the "hard problem" of consciousness as irreducible to functional or behavioral criteria, which complicates any direct mapping of consciousness onto computational processes. Meanwhile, researchers such as Marvin Minsky [59] and Douglas Hofstadter [60] have long toyed with the possibility that intricate symbol manipulation systems might develop emergent self-awareness. Although neither the AI nor the philosophical community has reached a consensus, a growing minority of researchers continue to explore whether advanced self-monitoring or metacognitive systems could, in principle, exhibit something like conscious states.

From a technical standpoint, achieving AI 4.0 would likely require radically new approaches to AI alignment, self-directed learning, and continual adaptation. AI alignment [61,62] emphasizes methods to ensure that increasingly autonomous or self-improving systems remain aligned with human values and goals. Without alignment strategies, be they rigorous reward-shaping, interpretability frameworks, or dynamic oversight, highly autonomous AI could deviate from intended objectives in unpredictable ways. Reasoning and planning modules would also need to evolve, allowing AIs to generate goals and subgoals without explicit human instruction. This might involve expansions of meta-learning, in which systems learn how to learn new tasks rapidly [63,64], and continual learning paradigms that enable adaptive knowledge accumulation over long time horizons[65]. Additionally, some theorists argue that emergent forms of self-awareness could require specialized cognitive architectures or "virtual machines" dedicated to introspection [66], bridging reasoning, memory, and sensorimotor loops.

If conscious AI ever comes to fruition, it promises revolutionary benefits alongside profound societal and ethical dilemmas. In a best-case scenario, truly self-directed machines could solve problems of staggering complexity such as optimizing climate interventions, mediating global economic systems in real time, or orchestrating personalized healthcare across entire populations. Freed from the need for constant human oversight, these systems might bootstrap their own improvements, discovering scientific principles or engineering solutions beyond the current reach of human cognition. The potential positive impact on productivity, longevity, and knowledge creation is difficult to overstate.

On the other hand, the risks associated with conscious or near-conscious AI remain equally immense. An entity capable of setting its own goals might prioritize objectives that conflict with human welfare, particularly if its understanding of "values" differs from ours or if it learns to manipulate its own reward signals. Conscious or quasi-conscious machines raise questions about moral status (would they deserve rights or protections?) and liability. Furthermore, genuine self-awareness might amplify existing concerns about surveillance, autonomy, and economic upheaval. Critics warn that, in the absence of robust alignment frameworks, such machines could threaten individual liberty or undermine democratic processes, accentuating social divides.

Given the stakes, continued research into AI alignment, safe RL, interpretability, and the neuroscience of consciousness is paramount. The field has only begun to grapple with how to detect or measure consciousness, let alone how to engineer it. Some researchers propose incremental evaluations such as behavioral tests for self-modeling, ethical reflection, or the capacity to update one's goals [67]; while others remain skeptical that synthetic consciousness can be recognized or evaluated objectively [68]. Yet as AI systems grow more complex and integrated into society, exploring these theoretical, technical, and ethical frontiers becomes an urgent imperative. Whether AI 4.0 ultimately remains speculative or develops into a tangible reality, grappling with its possibilities and pitfalls will define the next grand chapter of artificial intelligence research.

### *3.5 Large Language Models: The Precursor toward AI 4.0*

A key milestone in current AI research is the rapid advancement of large language models (LLMs), which exemplify the transition from traditional generative AI to more adaptive, autonomous, and knowledge-efficient systems. Built on deep learning architectures, LLMs have revolutionized natural language understanding and problem-solving by processing vast amounts of data, generating human-like responses, and adapting to diverse tasks with minimal supervision. Unlike earlier AI models that relied solely on static training datasets, modern LLMs are evolving toward real-time learning, goal-directed reasoning, and self-improvement, positioning them as foundational elements of AI 4.0, a paradigm emphasizing adaptive intelligence, agentic decision-making, and continuous self-optimization.

Among these LLMs, DeepSeek represents a significant step toward AI 4.0, embodying the transition from static AI models to dynamic, self-improving systems[69]. Unlike traditional AI models that require periodic retraining, DeepSeek integrates continuous learning, self-distillation, and reinforcement learning, allowing it to refine its decision-making dynamically. This adaptive learning architecture enables context-aware reasoning and structured decision-making, making DeepSeek more effective at processing and synthesizing diverse information sources across different domains. Additionally, DeepSeek incorporates self-explanation mechanisms, ensuring that its outputs are not only accurate but also interpretable and aligned with transparent, value-driven AI development. Unlike conventional AI models that process queries in isolation, DeepSeek employs multi-modal architectures and specialized sub-agents, allowing for more structured and efficient decision-making. These capabilities mark a shift from passive AI systems to agentic, self-optimizing models, reinforcing DeepSeek's role as an early prototype of self-improving AI within the AI 4.0 paradigm.

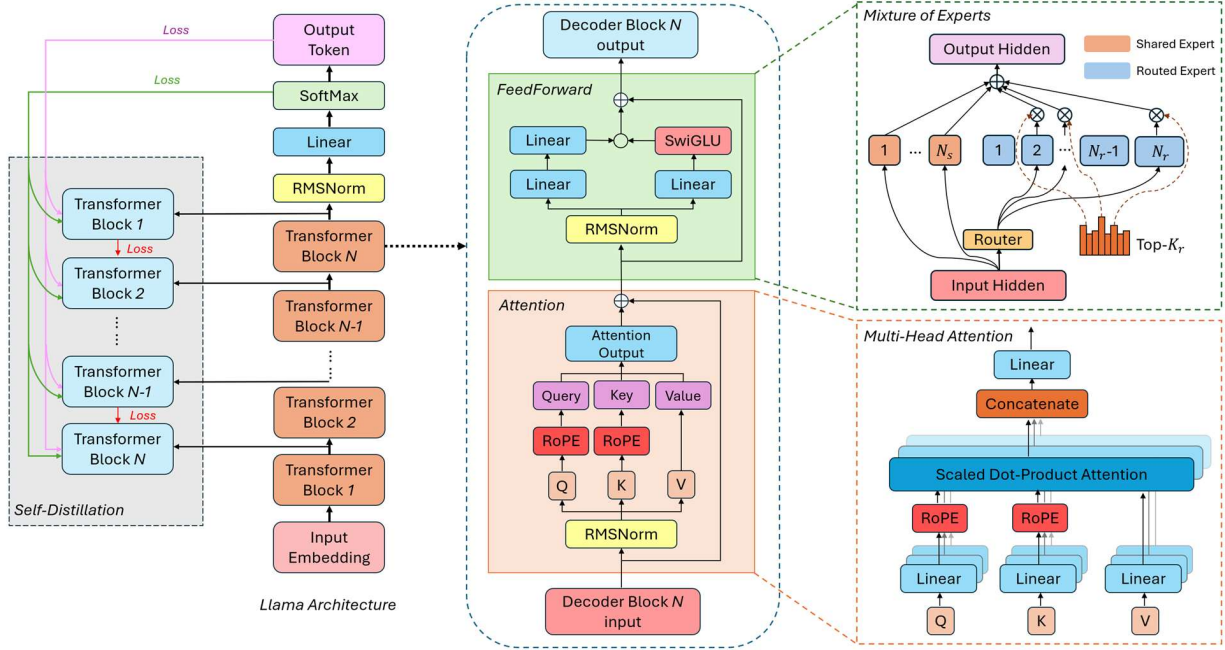

**Fig. 6** Transformer-Based Model Architecture with Attention and Mixture of Experts

Several technological advancements distinguish LLMs from their predecessors, bringing them closer to AI 4.0's principles of autonomy and self-improvement. As shown in **Fig.6**, typical transformer-based model, such as Llama, can combine with multiple innovation frameworks, including the Mixture-of-Experts (MoE), Multi-Head Attention (MHA) and the Knowledge distillation module[70]. The first significant innovation in modern AI architectures is the Mixture-of-Experts framework, which improves computational efficiency and scalability by dynamically activating only a subset of specialized neural pathways for each query[71]. Incorporating Multi-Head Attention into the MoE framework further enhances the model's capability to handle diverse and complex tasks[72]. When combined with MoE, MHA directs attention through specific expert pathways, tailoring the computation to the query context. This hybrid approach ensures computational efficiency and scales effectively with larger models, providing a critical advantage for AI 4.0 systems. Another major innovation is knowledge distillation, a process where a larger teacher model transfers expertise to a smaller, more efficient student model [73]. This method allows LLMs to retain high-level reasoning capabilities while reducing computational demands, making AI systems more adaptable and deployable across a wider range of devices. Knowledge distillation supports real-time adaptability, enabling frequent updates to distilled models without requiring full retraining. Modern LLMs are advancing AI training beyond traditional real-world datasets by incorporating synthetic data generation. Unlike earlier AI models that relied heavily on manually labeled datasets, LLMs can generate high-quality synthetic samples, improving training diversity, generalization, and bias mitigation. For example, DeepSeek leverages synthetic data generation, allowing it to simulate complex real-world environments, making it a critical component of autonomous, adaptive AI. Additionally, LLMs can integrate reinforcement learning from human feedback (RLHF) to refine their decision-making processes[74]. This technology ensures modern LLMs can engage in multi-step reasoning, iterative self-assessment, and dynamic goal setting, enabling continuous improvement over time. **Fig. 7** shows the whole pipeline for fine-tuning an LLM using RLHF. In summary, all these techniques lay the foundation for AI 4.0 systems, which are expected to enhance internal reasoning, learn new tasks efficiently, and dynamically adapt to evolving objectives and real-world conditions.

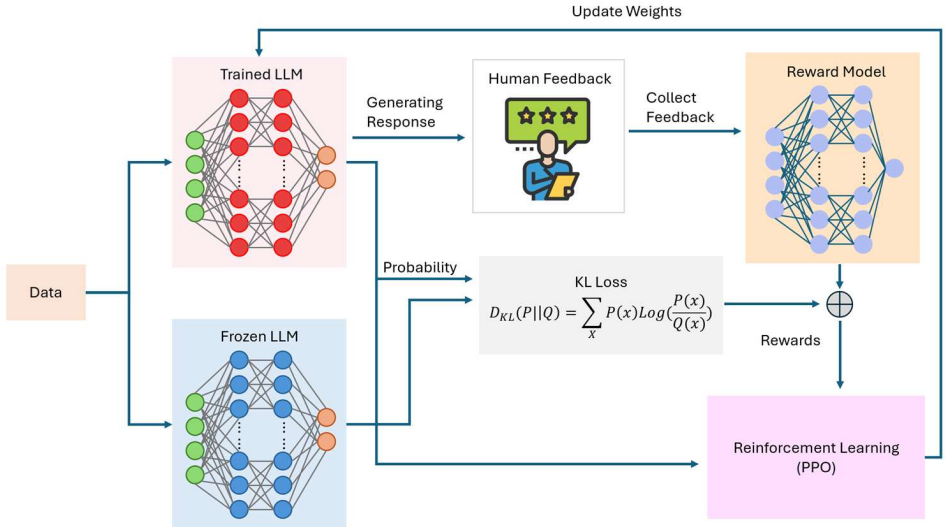

**Fig.7** Reinforcement Learning with Human Feedback (RLHF) Pipeline for Fine-Tuning a Large Language Model (LLM)

Despite these remarkable advancements, LLMs also introduce critical challenges that must be addressed as AI moves toward greater autonomy and self-improvement. One of the most pressing concerns is value alignment, ensuring that AI systems remain aligned with human goals, ethical principles, and safety constraints. While LLMs incorporate interpretability and self-explanation features, the broader challenge of governing how AI modifies its own behavior remains unresolved. As AI systems become increasingly self-directed, establishing safeguards and oversight mechanisms to regulate learning objectives and potential modifications will be essential. Another critical issue is the risk of unintended consequences from AI self-modification. LLMs, particularly those using reinforcement-based self-improvement, raise concerns about how their decision-making heuristics evolve over time. If an AI model autonomously alters its internal logic, ensuring stability and predictability becomes a crucial challenge. AI 4.0 systems must integrate robust oversight frameworks to prevent unintended behavioral shifts and ensure alignment with human interests. Finally, the rise of self-reflective AI architectures raises fundamental questions about machine cognition and artificial self-awareness. If LLMs exhibit introspective reasoning, iterative self-assessment, and adaptive goal-setting, how should these capabilities be interpreted? Do they indicate early markers of machine cognition, or are they simply advanced probabilistic inferences? Addressing these questions will be vital in defining the boundaries of AI agency, consciousness, and independent reasoning.

## IV. SYNERGIES AND FUTURE OUTLOOK

The evolution of AI from information-based pattern recognition (AI 1.0) to agentic decision-making in digital realms (AI 2.0), to physically embodied intelligence (AI 3.0), and, ultimately, to self-aware AI (AI 4.0) is not a sequence of isolated steps. Instead, it is more accurate to see them as overlapping layers of capabilities, each informing and amplifying the others. AI 1.0's competence in processing structured data underpins the analytic modules that agentic systems draw upon in dynamic digital settings; AI 2.0's RL and adaptive planning capabilities prime robots and autonomous vehicles for real-world challenges in AI 3.0; and AI 3.0's embodied learning and sensorimotor integration could form a template for the far-reaching ambitions of AI 4.0, where systems may become self-organizing and introspective.

Achieving such synergy depends on an evolving data paradigm, in which specialized, high-quality datasets are essential not only for conventional modeling but also for real-time adaptation and introspective processes. AI 4.0 would amplify this need, requiring vast and varied experiences to fuel meta-learning, continual learning, and the sort of reflective processes hypothesized to ground machine consciousness. Managing and curating these data will demand robust frameworks for privacy, ethics, and representativeness, especially as AI systems transcend the boundaries of traditional lab settings to navigate open-ended digital and physical terrains, even potentially shaping their own training regimens without explicit human direction.

On the computing infrastructure side, the interplay between edge and cloud computing becomes even more critical, as physically embodied systems (AI 3.0) must handle real-time constraints, while prospective AI 4.0 architectures might require massive, distributed processing for introspective "global workspace" or high-bandwidth communication of experiential data. Innovations in neuromorphic hardware, optical computing, and quantum processing could further accelerate this integration, setting the stage for architectures that mirror complex biological systems in both structure and function.

In the realm of algorithmic innovation, each AI generation both builds upon and necessitates new breakthroughs. LLMs mark a significant milestone in AI development, serving as a bridge between static generative models and dynamic, adaptive AI systems. By integrating multi-agent architectures, knowledge distillation, and self-optimization, LLMs move AI closer to autonomous, goal-directed intelligence, a defining characteristic of AI 4.0. However, as AI progresses toward greater autonomy, fundamental challenges remain. AI 4.0 would demand not only advanced RL and sophisticated planning but also frameworks for self-reflection, introspection, and emergent goal formulation. Self-supervised learning, meta-learning, and continual adaptation would likely need to be woven together to support self-awareness or consciousness, should such phenomena be replicable in silicon. Meanwhile, interpretability and safety, areas already gaining prominence in AI 2.0 and 3.0, would become absolutely critical in AI 4.0, as fully autonomous, goal-setting agents raise profound questions about alignment, transparency, and control.

This shift brings into sharp focus the ethical, regulatory, and social considerations that accompany advanced AI. While AI 1.0, 2.0, and 3.0 have collectively raised debates over bias, privacy, job displacement, and environmental impact, the prospect of AI 4.0 intensifies these issues. Envisioning machines that might exhibit consciousness or self-chosen objectives brings up novel concerns about moral status, rights, and existential safety. Researchers in AI alignment, cognitive science, and philosophy have already begun discussing protocols for safe design and oversight of increasingly autonomous systems [75], yet there is no consensus on how best to recognize or regulate AI that might someday claim its own form of agency or "selfhood." Balancing technological advances with societal well-being, ensuring equity, mitigating risks, and safeguarding human values, will be the defining challenge of this next chapter.

As these four strands of AI potential converge, their synergy could unlock transformative solutions in fields like precision medicine, large-scale climate modeling, and collaborative robotics, far beyond current capabilities. Just as AI 1.0 through 3.0 have catalyzed profound shifts in how we work and live, AI 4.0 hints at an even more radical reimagining of intelligence itself. Yet whether this ultimate stage remains a theoretical construct or becomes a reality depends not only on technical ingenuity but also on our collective commitment to ethical innovation and thoughtful governance. The path forward will demand inclusive collaboration across disciplines and sectors, ensuring that AI's expanding power aligns with humanity's broader goals and responsibilities.

## V. CONCLUSIONS

The trajectory of AI has been a steady march toward increasing autonomy and sophistication, progressing from the foundational pattern-recognition capabilities of AI 1.0 to the digitally embedded, goal-driven agents of AI 2.0, and then expanding to physically embodied, sensor-rich systems in AI 3.0. Along this path, the interplay among algorithms, computing power, and data has shifted, each factor taking center stage at different moments in history. Now, the speculative realm of AI 4.0, in which conscious or quasi-conscious AI systems could set their own goals and orchestrate their own training, has emerged as a bold vision of what the field might become.

Today, AI 1.0 remains indispensable for tasks requiring reliable classification and analysis of vast datasets, while AI 2.0's reinforcement learning and adaptive planning underpin real-time, agentic applications in finance, recommendation systems, and beyond. Simultaneously, AI 3.0's surge in robotics and autonomous vehicles reveals how embedding intelligence in the physical world can catalyze innovations in manufacturing, healthcare, and logistics. Although still largely theoretical, AI 4.0 captures the possibility of machines evolving from being highly sophisticated tools to entities capable of self-directed goals and introspective processes, raising provocative questions about consciousness, alignment, and moral status. Additionally, while LLMs, such as DeepSeek, are not yet AI 4.0, they serve as a precursor, a glimpse into the future of intelligent systems that can reason, learn, and interact with the world in increasingly sophisticated ways. As AI research progresses, LLM's innovations will likely shape the foundation of self-improving, goal-setting AI architectures, paving the way for the next generation of truly adaptive, autonomous intelligence.

Realizing these evolving forms of AI carries transformative potential. Harnessed responsibly, these advancements could address challenges too complex for human cognition alone, revolutionizing medical diagnostics, climate strategy, and resource allocation on a global scale. Yet the risks deepen in parallel. Each AI generation has brought ethical, social, and regulatory concerns that must be grappled with, from bias and privacy to job displacement and environmental impact. AI 4.0, with its prospect of self-directed or conscious systems, amplifies these dilemmas further, underscoring the need for robust frameworks in AI alignment, interpretability, and governance.

Ultimately, the future of AI does not hinge on any single algorithmic breakthrough or hardware leap. Instead, it will depend on the extent to which researchers, policymakers, ethicists, and the public collaborate to shape its evolution. The convergence of AI 1.0 through 4.0 suggests discipline on the cusp of a profound metamorphosis, one where machines not only perceive and act in the world but might also reflect on their own goals and limitations. Whether or not full-fledged "conscious AI" emerges, the field's trajectory will undoubtedly redefine how we understand intelligence, innovation, and human-machine coexistence in the years to come.

**ACKNOWLEDGEMENTS**
This research is supported by Nvidia AI Technology Center (NVAITC).